\documentclass{IOS-Book-Article}
\usepackage{enumitem} 
\usepackage{mathptmx}
\usepackage{graphicx}       
\usepackage{amsmath,latexsym,amsbsy,amssymb}
\usepackage{color}

\newtheorem{Lemma}{Lemma}[section] 
\newtheorem{Algorithm}{Algorithm}[section]

\def\hb{\hbox to 10.7 cm{}}

\begin{document}

\pagestyle{headings}
\def\thepage{}

\begin{frontmatter}              % The preamble begins here.

%\pretitle{Pretitle}
\title{OASIS: An Active Framework for Set Inversion}

%\markboth{}{March 24, 2018\hb}
%\subtitle{Subtitle}

\author[A,D]{\fnms{Binh T.} \snm{Nguyen}%
\thanks{Corresponding Author: Binh T. Nguyen (e-mail:
ngtbinh@hcmus.edu.vn). This is a preprint version of the main paper.}},
\author[A]{\fnms{Duy M.} \snm{Nguyen}},
\author[B]{\fnms{Lam Si Tung} \snm{Ho}},
\author[C]{\fnms{Vu} \snm{Dinh}}

%\runningauthor{Binh T. Nguyen et al.}
\address[A]{VNU HCM University of Science, Vietnam}
\address[B]{Dalhousie University, Halifax, Nova Scotia, Canada}
\address[C]{University of Delaware, USA}
\address[D]{Inspectorio Research Lab, Vietnam}

\begin{abstract}
In this work, we introduce a novel method for solving the set inversion problem by formulating it as a binary classification problem. 
Aiming to develop a fast algorithm that can work effectively with high-dimensional and computationally expensive nonlinear models, we focus on active learning, a family of new and powerful techniques which can achieve the same level of accuracy with fewer data points compared to traditional learning methods. 
Specifically, we propose OASIS, an active learning framework using Support Vector Machine algorithms for solving the problem of set inversion.
Our method works well in high dimensions and its computational cost is relatively robust to the increase of dimension. 
We illustrate the performance of OASIS by several simulation studies and show that our algorithm outperforms VISIA, the state-of-the-art method.
\end{abstract}

\end{frontmatter}
%\markboth{Binh et al.\hb}{Binh et al.\hb}
%\thispagestyle{empty}
%\pagestyle{empty}

\section{Introduction}
\begin{figure*}[ht]
	\centering
	\includegraphics[width=0.75\linewidth]{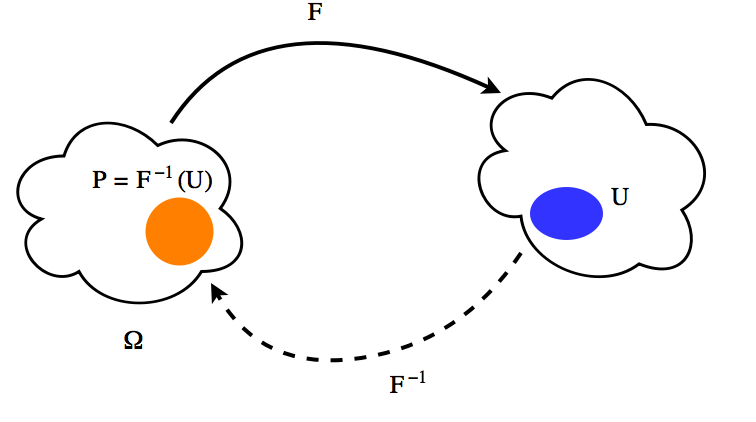}
	\caption{A set inversion problem with a given smooth function $F$ on a compact subset $\Omega$ in $\mathbb{R}^s$.}
	\label{fig:set_inversion_definition}
\end{figure*}

%During the last decades, set inversion methods have been extensively studied and gained a lot of interests in different research fields and applications. 
Set inversion methods have been successfully applied to many applications in sciences and engineering in the past decades.
Those include non-convex optimization problems \cite{Hammer95,Kearfott96,moore79}, nonlinear parameter set estimation \cite{kieffer1998interval}, localization and characterization of stability domains of dynamical systems \cite{colle2012mobile,walter1994guaranteed,drevelle2011set}, fault detection and identification \cite{jauberthie2012fault}, set-membership experimental design of biological systems \cite{marvel2012set}, and behaviour discrimination of enzymatic reaction networks \cite{donze2009,dinh2014effective}.
The problem of set inversion is formalized as follows:
given a set $U \subset \mathbb{R}^t$ and a smooth function $F: \Omega  \to \mathbb{R}^t$ where the state space $\Omega$ is a compact subset of $\mathbb{R}^s$, we want to determine the pre-image $P = F^{-1}(U)$ (Figure \ref{fig:set_inversion_definition}).
Despite the simplicity of its formulation, this problem is challenging due to the fact that the computation of the pre-image $P$ can be expensive because of the complicated geometric structure of $P$ and the cost to evaluate the forward function $F$.

Over the years, several computational methods related to set inversion have been proposed, most of which take advantage on expert's knowledges about the geometric structures of the pre-image. For linear systems, the solution set is a convex polyhedron for which an exact characterization can be attempted if the dimensions of the problem are not too large. Other simple-shaped forms, such as ellipsoids \cite{lesecq2003numerical,milanese1991estimation}, parallelotopes, zonotopes \cite{alamo2005guaranteed} or boxes have also been used to give an enclosure of the exact solution set. 

For non-linear models, the most popular approach is interval analysis, in which one can represent real numbers or integers by intervals and real vectors by boxes containing them. In this context, an interval $\left[{x}\right]$ can be described as a closed and connected subset in the set of real numbers $\mathbb{R}$:
\begin{equation*}
\left[{x}\right] =\left\{{x \in \mathbb{R} \ \ | \ \ \ \ a\le x\le b, \ \ \ \ a,b \in \mathbb{R}}\right\},
\end{equation*}
in which $a$ and $b$ are respectively the lower bound and the upper bound of $\left[{x}\right]$, the width of the interval $\left[{x}\right]$ is
$
w([x]) = b-a,
$
and its center is 
$
\text{center}([x]) = (b+a)/2.
$
%Especially, when $a=b$, $[x]$ is ``degenerate''. It turns out that 
Any real number can be considered as a degenerate interval in $\mathbb{R}$.
In a high-dimension space, a box $\left[{\bf{x}}\right]$ in $\mathbb{R}^n$ is a Cartesian product of $n$ real intervals:
$
\left[{\bf{x}}\right] = \times_{i=1}^{n} \left[{x_i}\right],
$
where $\left[{x_i}\right]$ is an interval in $\mathbb{R}$. 
By these definitions , it is quite natural to extend all classical real arithmetic operations and elementary functions to intervals. A subpaving of a box $\left[{\bf{x}}\right]$ in $\mathbb{R}^n$ can be defined as the union of non-overlapping sub-boxes of $\left[{\bf{x}}\right]$ with non-zero width. 
%It is interesting to emphasize that for any compact set $U$, one can compute a guaranteed approximation of $U$ by an inner subpaving $U^-$ and an outer subpaving $U^+$ such that $U^- \subset U \subset U^+$.

To solve the set inversion problem, Jaulin and Walter characterize the solution set of a system of non-linear real constraints by enclosing it between internal and external unions of interval boxes as mentioned above.
To be more specific, they use an inclusion function of $F$ for approximating two regular subpavings $P^-$ and $P^+$ such that
$
P^- \subset P \subset P^+
$
by the algorithm SIVIA (Set Inverter Via Interval Analysis) \cite{jaulin1993set}. 
For a given forward function $F$, its inclusion function $[F]$ is defined as
\begin{equation*}
[F]([{\bf x}]) = \left[{\left\{{F({\bf x}) | {\bf x} \in[{\bf x}] }\right\}}\right].
\end{equation*}
In other words, $[F]([{\bf x}])$ is the smallest box in $\mathbb{R}^t$ that contains $F([{\bf x}])$.
SIVIA  can be applied for any function $F$ whose an inclusion function can be calculated and it works well with non-linear forward models in spite of complicated set geometries.
However, this algorithm relies heavily on sub-partitioning covering elements to approximate the pre-image. 
As a results, the computational cost increases exponentially in terms of the dimension, rendering the method impractical for studies of medium and high-dimensional functions.

There have been lots of useful applications using machine learning techniques in various aspects of daily life. Duy et al. use deep neural networks and Gaussian mixture models for extracting brain tissues including gray matter, white matter, and cerebrospinal fluid from high-resolution magnetic resonance images \cite{MinhDuy2017}. Cuong et al. propose a novel approach by using Mel-frequency cepstral coefficients (MFCCs) for deriving different types of features and solving the eye movement identification with a multi-class classification approach  \cite{6495054}. Deep neural networks have been applied to handwritten characters recognition  \cite{Simard2003} and food recognition \cite{indeed18}. Recently, there has been a new trend  in applied mathematics using a machine learning approach for solving non-linear \cite{RAISSI2018125} and linear partial differential equations \cite{RAISSI2017683}. 

In this paper, we take a different approach to the problem of set inversion. The main idea is to formulate the task of reconstructing the pre-image as classifying points on the state space into two subsets: those lie inside the pre-image, and those do not. 
Instead of using interval analysis to verify if certain set is a subset of the pre-image, we use a sampling scheme to sample a sequence of inputs from the state space and check if the points in the sequence belong to the pre-image simply by evaluating the forward function. 
Based on those samples, we employ a machine learning algorithm to compute the pre-image. 
Specifically, we introduce an active learning Support Vector Machine (SVM) algorithm for set inversion which is implemented in a Python-package OASIS (Optimization-based Active learning for Set Inversion with Support vector machine).
%As our knowledge, this is the first time machine learning is applied for the set inversion problem.
%The new approach enables the use of various machine learning algorithms to solve the problem of set inversion. 
Our method can work efficiently with high-dimensional and computationally expensive models because its active learning step aggressively searches for high-informational samples.
Therefore, OASIS requires less data points for achieving the same level of accuracy compared to traditional methods.
%, a family of powerful supervised learning protocols capable of producing more accurate classifiers while using a smaller number of labeled data points than traditional learning methods. 
%The computational cost of our method is relatively insensitive to the increasing of dimensionality.
%The paper is organized as follows. In Section 2, we introduce an active learning framework for set inversion and propose a fast randomized active support vector machine algorithm. We then investigate the problem in different examples to demonstrate its efficacy in Section 3. Finally, we conclude the paper by some discussions and description of future works.
Additionally, our method inherits two important qualities from Support Vector Machine: working well with high dimension and having fast prediction time.
Our simulation studies demonstrate that OASIS outperforms SIVIA on these aspects. 

\section{OASIS: Optimization-based Active learning framework for Set Inversion with Support vector machine}

In this section, we first describe our proposed approach and then explain how to implement the corresponding algorithm for solving the set inversion problem under an active learning framework.

\subsection{Classification: a machine learning approach to set inversion}

Classification is the problem of identifying to which of a set of categories a new observation belongs, on the basis of a training set of data containing observations whose category membership is known. A mathematical function mapping input data to categories is known as a classifier.

We define a classifier $\phi$ that assigns every point $x$ in the state space $\Omega$ a label of either $1$ or $-1$ as follows:
\begin{equation}
\phi(x)=
\left\{
	\begin{array}{ll}
		1  & \mbox{if } ~F(x) \in U \\
		-1 & \mbox{if } ~F(x) \not \in U
	\end{array}
\right.
\end{equation}
Thus, the problem of set inversion can be regarded as a classification which we want to learn about $\phi$.
Since the closed form of $\phi$ is unknown and evaluating $\phi$ might be expensive, we want to approximate $\phi$ by a simple classifier $h$ with small classification error.

In this paper, we use Support Vector Machine (SVM) \cite{NIPS1996_1187}, a popular supervised learning method, to learn $\phi$. 
Suppose that a training sequence of inputs $\{x_i\}_{i=1}^N$ is generated and their classification $y_i=\phi(x_i)$ is achieved, the algorithm seeks for a manifold of the form $\psi(  x)=0$ that separates the samples of label $-1$ from those with label $1$,  where $\psi(  x)$ can be written as
\begin{equation}
\psi(  x)=b+\sum_{k=1}^{N} \beta_k y_k\mathcal{K}(  x_k,   x).
\label{GRBF}
\end{equation}
Here $\mathcal{K}(\cdot,\cdot)$ is the so-called kernel function and $\beta_k$ denotes the $k$th coefficient of the SVM decision function. The coefficients of the separating manifold are obtained by solving the following constrained optimization problem:
\begin{align}
\nonumber   \beta_{SVM}=\arg\max & \sum_{k=1}^{N} \beta_k - \frac{1}{2} \sum_{k=1}^{N} \sum_{j=1}^{N} \beta_k\beta_j y_k y_j \mathcal{K}(  x_k,   x_j), \\
\text{subject to: } & \sum_{k=1}^{N} \beta_k y_k=0, \label{eq:dual_svm_nonlin} \\
\nonumber & 0\leq \beta_k\leq L,\,\,\, \forall\,\, k=1,2,\cdots,N.
\end{align}

Throughout this work, we focus on a commonly used Gaussian Radial Basis Function (GRBF) kernel function with variance $\gamma^2$, which is of the form:
\begin{equation}
\mathcal{K}_{RBF}(x,y)=\exp\left(-\frac{\|  x-y\|^2}{2\gamma^2}\right).
\label{eq:g_krnl}
\end{equation}
%where the kernel variance $\gamma$ is a design variable. 
We have the following result about universality of GRBF.

\begin{Lemma}[\cite{steinwart2002influence}]
The GRBF kernel can separate any disjoint pair of finite subsets $K_1$ and $K_{-1}$ of $\mathbb{R}^s$, that is, there exists $\psi$ of the form ($\ref{GRBF}$) with $N=|K_{1} \cup K_{-1}|$ such that 
\[
\psi(x)>0 ~ \forall x \in K_1 ~~ \text{and} ~~  \psi(x)<0 ~ \forall x \in K_{-1}.
\]
\label{svm}
\end{Lemma}

This lemma implies that for proper choice of kernel variance $\gamma$, the GRBF kernel is capable of separating any finite data set corresponding to their labels. We note that when data is separable by the kernel, the SVM decision function also produces a separating manifold.

\subsection{An active learning algorithm for set inversion}

Many decision-making processes can be formularized as supervised machine learning problems, for which a prediction function that maps an input to an output is constructed based on example input-output pairs. 
Traditionally, the training sequence of such a learning problem is generated independently from an unknown distribution. 
In many real-world applications, obtaining the labels can be quite costly and time consuming. This motivates the studies of active learning algorithms, a special class of supervised learning which aims at learning the prediction function while minimizing the number of labels requested. 

In contrast to the traditional passive setting, the queried sequence in an active setting is generated sequentially. 
An active learner takes into account information from previous samples to aggressively look for high-informational samples in an adaptive manner. 
Since the learner has the freedom to choose the training set, the number of examples for learning can be much smaller than the number required in a passive setting.

Our SVM-based active sampling scheme can be described simply by the following algorithm:

\begin{Algorithm}(OASIS)
\begin{enumerate}

\item {Input}: Starting a sequence $\{x_1,\ldots, x_k\}$ and the corresponding labels $\{y_1,\ldots, y_k\}$. 

\item{Initialize}: $n=k$

\item {While} $n<N$
\begin{itemize}
\item Configure $\gamma$ and  $\beta_{SVM}$ such that the empirical error of the classifier $\psi_n$ on the current sequence of samples is minimized.
\item Generate a random starting point $x_0$ on the state space $\Omega$
\item Find the nearest point ${x^*}$ to $x_0$ on the manifold $\psi_n(x)=0$ by solving the following optimization problem:
\begin{equation}
     \begin{aligned}
      \text{Minimize } & \left\|{x-x_0}\right\|_2  \\
      \text{Subject to} &\\
       &\psi_n(x)=0.
     \end{aligned}
     \phantom{\hspace{6cm}} %%<---adjust the value as you want
  \label{eqn:opt}
\end{equation}
\item $x_{n+1} = x^*$
\item $y_{n+1} = \phi(x_{n+1})$
\item $n=n+1$
\end{itemize}
\item {Output}: Sequence of active samples $\{x_1,\ldots, x_N\}$, $\{y_1,\ldots, y_N\}$ and the classifier $\psi_{N}$.
\end{enumerate}
\end{Algorithm}

%\begin{Algorithm}(Samp)
%\begin{enumerate}
%\item {Input}: Classifier $\psi(x)$. 
%\item Choose randomly a point ${x^*}$  in the manifold $\psi(x)=0$  by the Algorithm \ref{solving_manifold}.
%\item {Output}: Active sample ${x^*}$ 
%\end{enumerate}
%\end{Algorithm}

%\begin{Algorithm}(Random Selection)
%\begin{enumerate}
%\item {Input}: Classifier $\psi(x)$.
%\item Generate a random starting point $x_0$ on the state space $\Omega$.
%\item Find the nearest point ${x^*}$ to $x_0$ on the manifold $\psi(x)=0$ by solving the following optimization problem:
%\begin{equation}
%     \begin{aligned}
%      \text{Minimize } & \left\|{x-x_0}\right\|_2  \\
%      \text{Subject to} &\\
%       &\psi(x)=0.
%     \end{aligned}
%     \phantom{\hspace{6cm}} %%<---adjust the value as you want
%\end{equation}
%To do that, we use Sequential Quadratic Programming for obtaining the minimum value and the corresponding point  ${x^*}$ \cite{Nocedal2006}.
%\item {Output}: An active sample ${x^*}$.
%\end{enumerate}
%\label{solving_manifold}
%\end{Algorithm}

Note that, we use Sequential Quadratic Programming \cite{Nocedal2006} to solve the optimization problem \eqref{eqn:opt}.
At step 3, OASIS requires $\gamma$ to be calibrated so that the current samples are separated according to their labels, which is made possible by Lemma $\ref{svm}$. In practice, this step is done by starting with small values of $\gamma$ and keep increasing it until the  empirical classification error using SVM reduce to zero. The optimal value of $\gamma$ plays an important role in computing the guaranteed approximation of the pre-image in later part of the paper. 

We note that the active learning literature often divides active learning algorithms into two types: aggressive schemes and mellow schemes \cite{Dasgupta_2011}. 
This dichotomy comes from a simple exploration-exploitation trade-off: since active samplers focus on searching for high-informational samples to increase accuracy, they assume that the learning models and the collected labels are correct and often fails to recognize the existence of any model mismatch. 
Many active learning algorithms are thus implemented with some additional "exploration" components to alleviate this issue, with the cost that the algorithm would be less efficient. 

In the context of set inversion, the labels of the data are often assumed to be correct, making it a good context for an aggressive algorithm like OASIS. 
Our initializing phase by using a starting set of random data also acts as a safety net for the events that there are errors in training data. 
It also helps the algorithms to stabilize quickly and facilitates the convergence to the optimal prediction.

\section{Numerical results}

In this section, we evaluate the performance of our algorithm in various settings and make the comparison with VSIVIA V0.1 \cite{Herrero_anefficient}, a vector implementation of SIVIA algorithm in  Matlab. 
By employing these data structures, VSIVIA can optimize the algorithm and mitigate the recursion of a huge quantity of loops as well as function calls, therefore reduce obviously computation time compared to the original SIVIA \cite{Herrero_anefficient}.
%Our comparison consists of 3 distinct basic 2D shapes (circle, doughnut and ring), along with a sphere in higher dimensions (up to 8). 
In our simulation, we consider three distinct basic 2D shapes (circle, doughnut and ring) and spheres in higher dimensions (up to $8$) for the pre-image $P$.
To illustrate the application of our method, we utilize OASIS to study the predator-prey problem through the Lotka-Volterra model.
%We also run an empirical experiment about Lotka-Volterra problem showing that our algorithm can be well applied in other fields such as solving a system of ordinary differential equations. 
All experiments are performed on a computer with Intel(R) Core(TM) i7-5500U CPU running at 2.40GHz with 8GB of RAM.

\subsection{Comparison with VSIVIA}

In this comparison, we use OASIS and VSIVIA to reconstruct the pre-image corresponding to various forward functions.
For OASIS, we first randomly sample $100$ initial points and then use our active sampling procedure to obtain additional $400$ points.
The accuracy of OASIS is calculated by the fraction of correctly predicted points by the returned classifier on a testing set.
For VSIVIA , we use the default setting.
%In this comparison, beside accuracy which is calculated by dividing correctly predicted points by total points in a state space given, processing time is also taken into account to assess the efficacy of both algorithms. 
%Along with training time, which is a duration needed to construct a model, time needed to predict total points assigned for each set inversion is also considered. Table \ref{table:t1} and \ref{table:t2} offer a deep insight into the performances of two approaches, simultaneously, Figure \ref{fig:Compare} depicts the outcomes of 2D cases.

\subsubsection{Two-dimensional examples}

\begin{figure}
	\includegraphics[width=0.65 \linewidth]{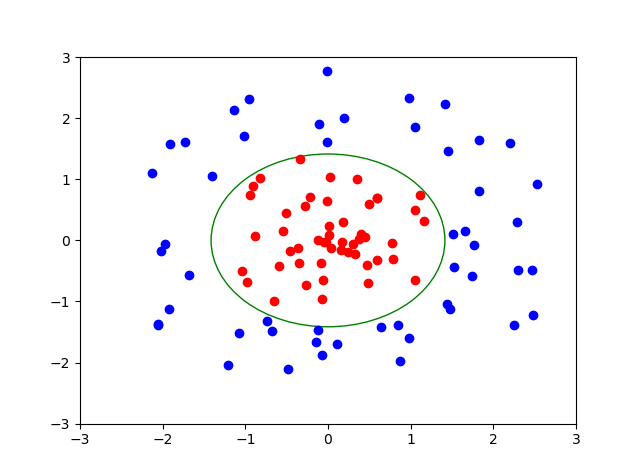}
	\caption{For recovering the boundary $\partial P$ of a region $P$ (i.e. a circle), one can do sampling an enough number of points with two labels: $1$ for all points inside the circle and $-1$ for the others. Finally, by using an appropriate classification model, one can approximately construct $\partial P$.} 
	\label{fig:2DSquare}
\end{figure}

\begin{figure}
	\includegraphics[width=1.05\linewidth]{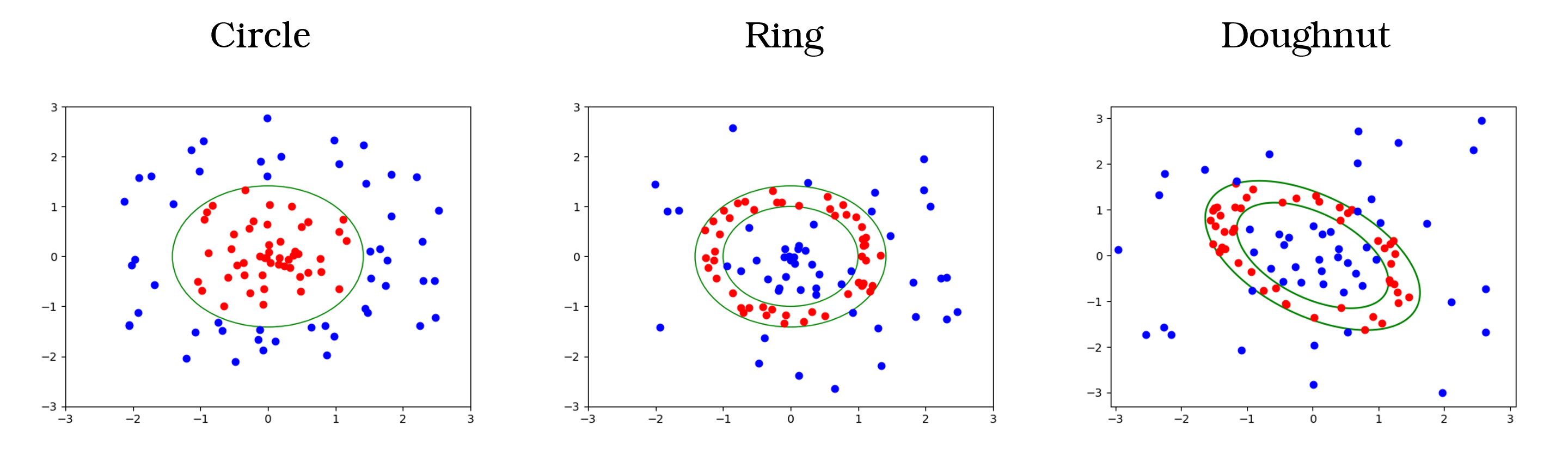}
	\caption{Distribution of randomly sampled points in $2$D shapes by OASIS.}
	\label{fig:2DRandom}
\end{figure}

For 2D examples, we consider two following forward functions:
\[
F_1(x,y) = x^2 + y^2 \quad \text{and} \quad F_2(x,y) = x^2 + y^2 + xy.
\]
Using these functions, we formularize three different set inversion problems: 
\begin{equation}
\begin{aligned}
P_{\text{circle}} &= F_1^{-1}([0,2]) = \{(x,y) \in \Omega_{2D} \mid x^2 + y^2 \leq 2\} \\
P_{\text{ring}} &= F_1^{-1}([1,2])  = \{(x,y) \in \Omega_{2D} \mid 1 \leq x^2 + y^2 \leq 2\} \\
P_{\text{doughnut}} &= F_2^{-1}([1,2]) = \{(x,y) \in \Omega_{2D} \mid 1 \leq x^2 + y^2 + xy\leq 2 \}
\end{aligned}
\label{eq:circle_ring_doughnut}
\end{equation}
where $\Omega_{2D} =  [-3, 3]\times[-3, 3]$ is the state space.
%The inversion of Circle, Ring and Doughnut are described by Eq. (\ref{eq:circle_ring_doughnut}), in which $R_1 = 1$, $R_2 = R = 2$ and the state space is $\Omega_{2D} = [-3, 3]\times[-3, 3]$. 
We recall that points in the pre-image are given label $1$ and points outside the pre-image are given label $-1$ (see Figure \ref{fig:2DSquare} for the visualization).
%For our algorithm, we initially random $k = 100$ points (50 for each class) then generate $N = 400$ points for active learning. 
The initial $100$ randomly sampled points is illustrated in Figure \ref{fig:2DRandom}.

\begin{figure}
	\includegraphics[width=1\linewidth]{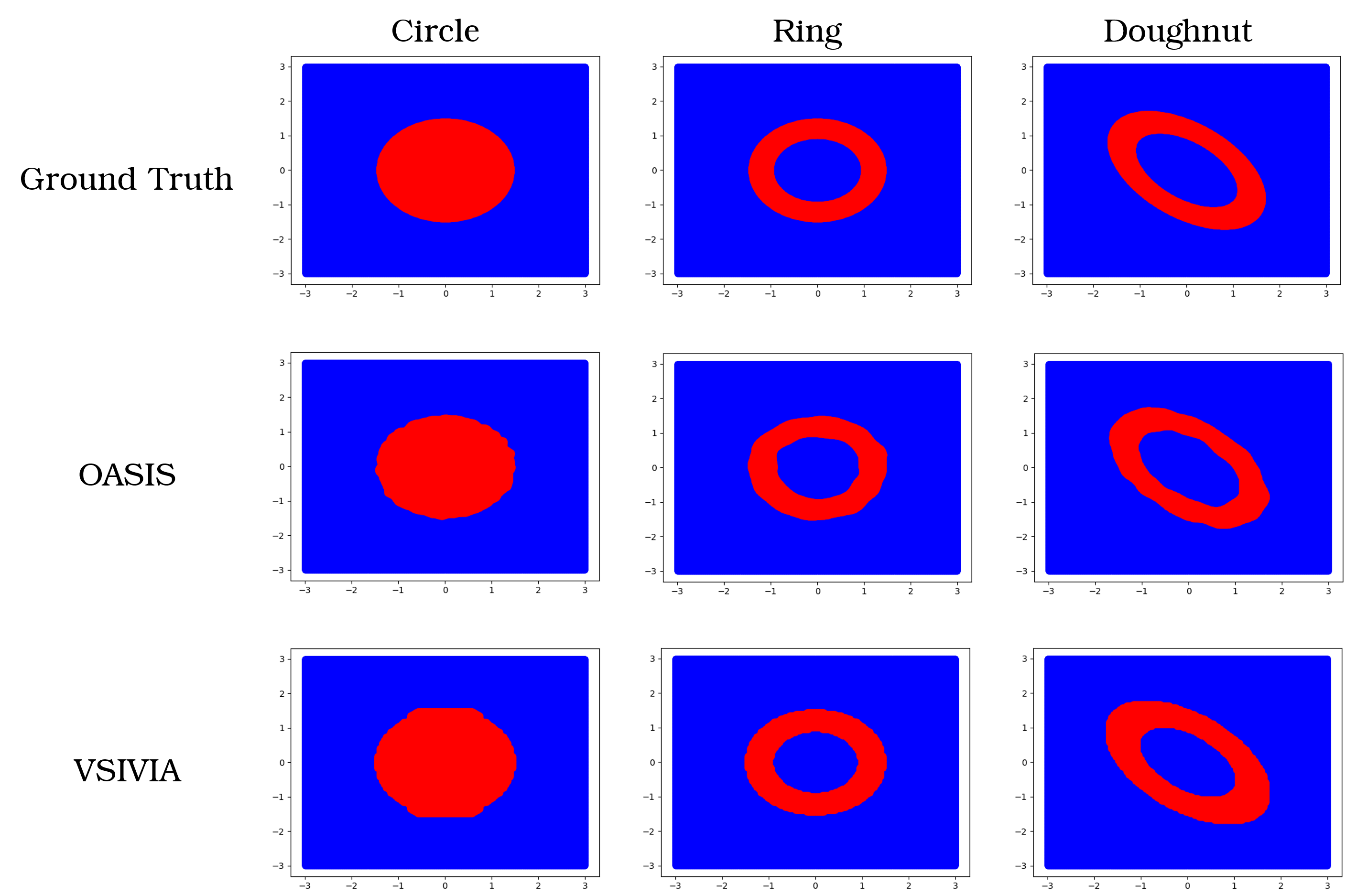}
	\caption{Outcomes produced by  OASIS  and VSIVIA compared to $2$D ground-truth shapes.}
	\label{fig:Compare}
\end{figure}

\begin{figure}
	\includegraphics[width=0.75\linewidth]{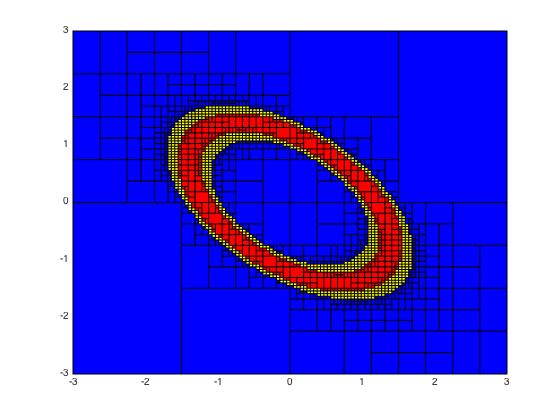}
	\caption{Graphical illustration of a result computed by VSIVIA in Doughnut example.}
	\label{fig:2DVSIVIA}
\end{figure}

To compare the accuracy of OASIS and VSIVIA, we construct an evenly spaced grid of $601$ points on each dimension and use them as the test set.
%scatter total 361201 points on the state space which need to be predicted. 
We note that VSIVIA doesn't explicitly provide a classifier for prediction.
This algorithm divides the space into certainty boxes, which are marked in blue if they lie outside of the pre-image, red if they are inside, and yellow if the algorithm is uncertain about their positions.
To do prediction for a data point using the output of VSIVIA, ones need to search for the box containing this point and make the decision based on the colour of the box (Figure \ref{fig:2DVSIVIA} indicates the result of VSIVIA for $P_{\text{doughnut}}$).
Throughout this study, we assume that the yellow region of VSIVIA lies inside the pre-image (Figure \ref{fig:Compare}).
Our method is slightly better at prediction compared to VSIVIA for all problems.
To be more specific, our average accuracy is $0.9935$ while the average accuracy of VSIVIA is $0.9843$.
    
As far as training time is concerned, VSIVIA is much faster than our algorithm for reconstruction of the pre-images in two dimensions.
The entire training time of VSIVIA algorithm is just roughly $0.4$ second while our method needs $2-4$ minutes since our method re-trains the classifier for each additional actively sampled point. On the other hand, our prediction time per point is noticeably faster than VSIVIA. 

We note that in many applications, the training period to obtain the classifier is a one-time cost that can be computed offline without significant time constraint, while the prediction of new data must be done as quickly and as accurately as possible. For example, in computer vision, the training period using deep learning may be allowed to run for weeks, while the prediction must be done in matter of seconds \cite{MinhDuy2017,Sun_2014}. 
Similarly, in the context of control theory, the offline-computed controller for decision-making is embedded into a device with limited memory and computational power \cite{bemporad2002explicit,chakrabarty2017support}. In such cases, our algorithm has the potential to make significant improvement over VSIVIA.

%However, our prediction time, which is less than 3 seconds for all shapes, is dramatically faster than VSIVIA with the average of 11.33 seconds. 

\begin{center}
\begin{table}    
    \begin{tabular}{c c c c c} 
        \hline
        \textbf{Shapes} & \textbf{Accuracy} & \textbf{Training Time (s)} & \textbf{Average Prediction Time (ms)} & \textbf{Resolution}\\ 
        \hline
        Circle & 99.61\% & 124.88 & 0.00789 & 601 \\
        \hline
        Doughnut & 99.18\% & 221.83 &  0.00811  & 601\\
        \hline
        Ring & 99.28\% & 71.76 & 0.00825  & 601\\
        \hline
        Sphere 3D & 99.34\% & 81.22 & 0.00828 & 151\\
        \hline
        Sphere 4D & 99.73\% & 90.95 & 0.00849  & 41\\
        \hline
        Sphere 5D & 99.63\% & 132.51 & 0.00892 & 31\\
        \hline
        Sphere 6D & 99.86\% & 174.64 & 0.01043 & 18\\ 
        \hline
        Sphere 7D & 99.96\% & 372.01 & 0.01348 & 12\\
        \hline
        Sphere 8D & 99.98\% & 452.26 & 0.01245 & 9\\
        \hline
    \end{tabular}
    \caption{Performance of OASIS algorithm.}
    \label{table:t1}
\end{table}
\end{center}

\begin{center}
\begin{table}
    \begin{tabular}{c c c c} 
        \hline
        \textbf{Shapes} & \textbf{Accuracy} & \textbf{Training Time (s)} & \textbf{Average Prediction Time (ms)}\\ 
        \hline
        Circle & 99.2\% & 0.3908 & 0.02395\\ 
        \hline
        Doughnut & 97.55\% & 0.4102 & 0.04482\\
        \hline
        Ring & 98.73\% & 0.3553 & 0.02533\\
        \hline
        Sphere 3D & 99.11\% & 0.6828 & 0.03604\\
        \hline
        Sphere 4D & 99.58\% & 2.2174 & 0.25993\\
        \hline
        Sphere 5D & - & - & - \\
        \hline
        Sphere 6D & - & - & - \\
        \hline
        Sphere 7D & - & - & - \\
        \hline
        Sphere 8D & - & - & - \\
        \hline
    \end{tabular}
    \caption{Performance of VSIVIA algorithm.}
    \label{table:t2}
\end{table}
\end{center}
%\begin{center}
%	\begin{table}
%		\begin{tabular}{c c c c} 
%			\hline
%			\textbf{Shapes} & \textbf{Accuracy} & \textbf{Training Time (s)} & \textbf{Prediction Time (s)}\\ 
%			\hline
%			Sphere 6D & 0.9986 & 174.64 & 354.68\\ 
%			\hline
%			Sphere 7D & 0.9996 & 372.01 & 483.14\\
%			\hline
%			Sphere 8D & 0.9998 & 452.26 & 536.08\\
%			\hline
%		\end{tabular}
%		\caption{Performance of our algorithm for Sphere in much more higher dimension.}
%		\label{table:t3}
%	\end{table}
%\end{center}

\subsubsection{High dimensional problems}
The merit of our approach compared to VSIVIA is significantly reflected in high dimensions. 
To illustrate this, we consider problems where the pre-images are high-dimensional spheres.
Specifically, the spheres and the state spaces are defined as follows:
\begin{equation*}
\begin{aligned}
P_{3D} & = \{(x,y,z) \in \Omega_{3D} \mid x^2 + y^2 + z^2 \leq 0.5\} \\
P_{kD} & = \{(x_1, x_2, \ldots, x_k) \in \Omega_{kD} \mid \sum_{i=1}^k{x_i^2} \leq 0.25\} \quad k = 4, 5, 6, 7, 8 \\
\Omega_{3D} & = [-1.5, 1.5] \times [-1.5, 1.5] \times [-1.5, 1.5] \\
\Omega_{4D} &= [-1, 1] \times [-1, 1] \times [-1, 1] \times [-1, 1] \\
\Omega_{kD} &= [-0.75, 0.75]^k, \quad k = 5, 6, 7, 8.
\end{aligned}
\label{eq:sphere3D}
\end{equation*}
To compare the accuracy of OASIS and VSIVIA, we again construct an evenly spaced grid and use them as the test set.
The number of points on each dimension (resolution) for each problem is summarized in Table \ref{table:t1}.
%All parameters are set as similar as 2D shapes except the quantity of points need to be classified, which is 3,442,951 for 3D sphere while those of 4D and 5D are 2,825,761 and 28,629,151 respectively. 
Note that VSIVIA fails when the dimension is higher than four because of being out of memory.
Table \ref{table:t1} and Table \ref{table:t2} summarize the results of this simulation study.
For problems in 3D and 4D, we see that OASIS is slightly better than VSIVIA in prediction.
Similar to 2D cases, the training times of OASIS are slower than the training times of VSIVIA but the prediction times of OASIS is much faster.
%In term of processing time, since we perform with high dimensions, VSIVIA take longer time for training as it generates more boxes but it is still efficient with merely 0.68 second for 3D and 2.22 second for 4D. However, its prediction time enormously increases as numbers of points surge to around 3,000,000. Although the quantity to be predicted in 3D is more than 4D but the number of boxes generated by VSIVIA for the latter shape is significant higher than the former, therefore the time of prediction for Sphere 4D of this approach is extremely longer than that of 3D with 734.49 and 124.09 seconds, respectively. Our method still proves to be the prompt algorithm for predicting with the average of 26 seconds for both cases although our training time still significant higher than that of VSIVIA.

%We execute 3 more experiments of sphere in more higher dimensions, namely 6D, 7D and 8D, to proof that our algorithm can work well regardless of this factor. These problems are set up as similar as Sphere 5D with $R_{6D}=R_{7D}=R_{8D}=0.25$ and the space given for each axis is also $[-0.75, 0.75]$. In the light of our main purpose, which is solely to ensure that this method can run in high dimension, we divide discretely axises into 18, 12 and 9 points leading to the total points scattered in space given for each dimension are 34012224, 35831808 and 43046721, accordingly. Because of this way of scattering, the accuracy obtained for each experiment is amazingly high with almost 100$\%$. The numerical results of these set inversion issues are shown in Table \ref{table:t2}.

\subsection{Set inversion in predator-prey problem}

\begin{figure}
	\includegraphics[width=0.75\linewidth]{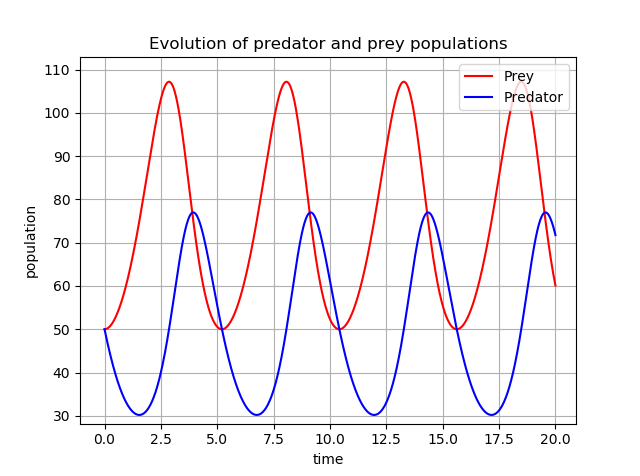}
	\caption{An example of a Lotka-Volterra model where $u(0) = v(0) = 50$, $p_1 = 1$, $p_2 = 0.02$, $p_3 = 1.5$, and $p_4 = 0.02$.}
	\label{fig:exampleLV}
\end{figure}

To show that our approach can be applied in practical applications, we consider the popular predator-prey problem through the Lotka-Volterra model. 
This model is represented by the following system of differential equations: 
\begin{align*}
\frac{du}{dt} &= u(p_1 - p_2v) \\
\frac{dv}{dt} &= -v(p_3 - p_4u),
\end{align*}
where $u, v$ denote the population of prey and predator, $p_1$ is the birth rate of the prey, $p_2$ is the dip in its number because of the predator, $p_3$ is the predator's death rate and $p_4$ is the growth of predator due to the interaction with prey. 
For example, Figure \ref{fig:exampleLV} depicts changes in the population of them with $u(0) = v(0) = 50, p_1 = 1, p_2 = 0.02, p_3 = 1.5$, and $p_4 = 0.02$.
    
\begin{figure}
	\includegraphics[width=0.75\linewidth]{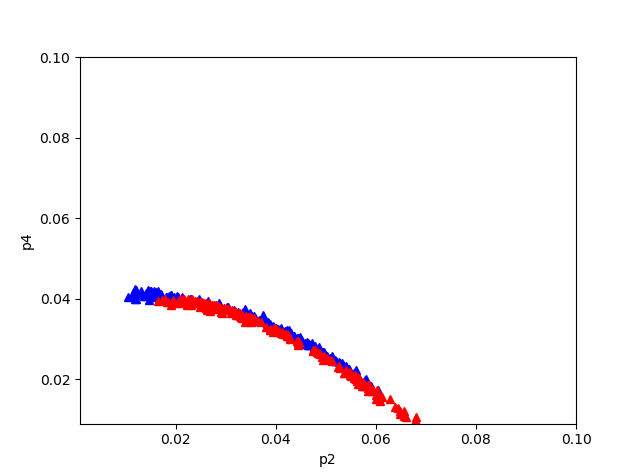}
	\caption{Illustration of 400 actively sampled points in the set inversion of Lotka-Voltera problem.}
	\label{fig:generateLV}
\end{figure}

\begin{figure}
	\includegraphics[width=1.05\linewidth]{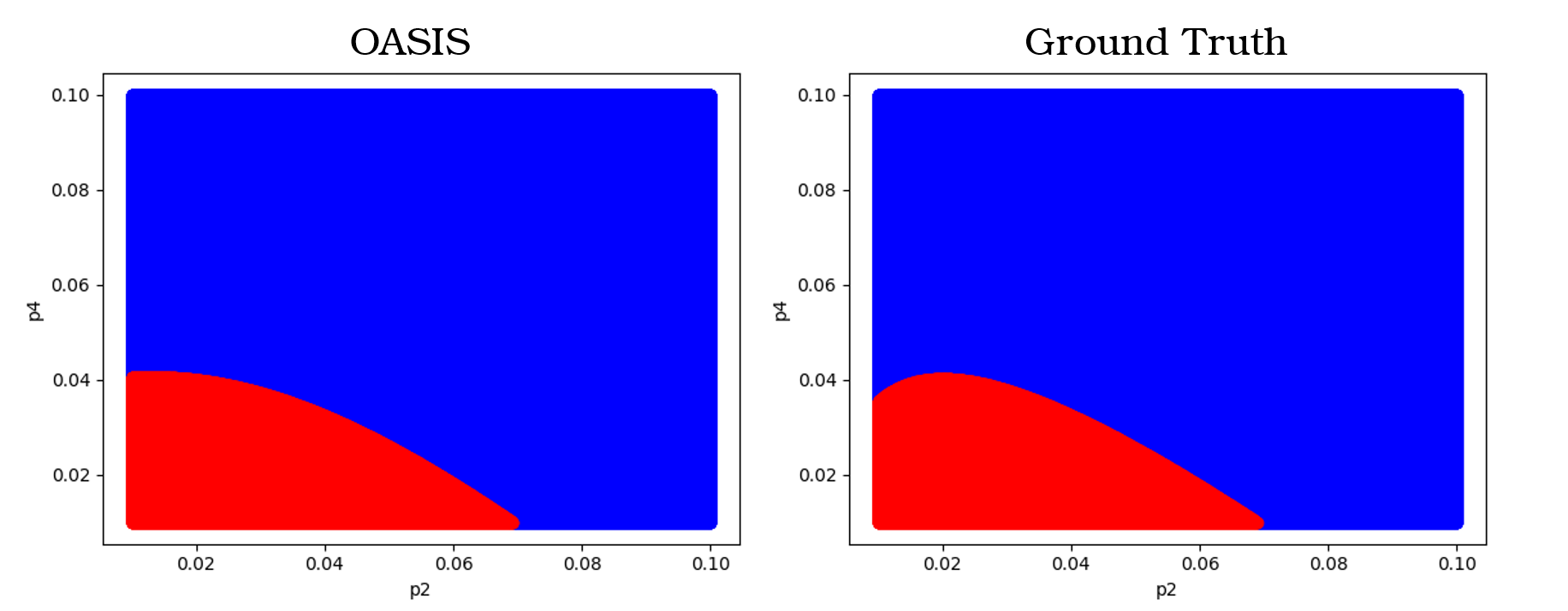}
	\caption{The result of OASIS algorithm for Lotka-Volterra problem compared to the ground truth.}
	\label{fig:resultLV}
\end{figure}

In this example, the set inversion problem is to construct the set of all parameters $(p_2, p_4)$ such that the minimum population of prey is always above $10$ throughout the period from $0$ to $20$, given that the initial populations of predator and prey are both $50$, the birth rate of the the prey is $1$, and the death rate of predator is also $1$.
The state space of this problem is $\Omega = [0.01, 0.1] \times [0.01, 0.1]$. 
%and total points need to be predicted is 1002001. 
To construct the classifier, OASIS randomly selects $400$ initial points then generate an additional of $400$ actively sampled points, which are illustrated in the Figure \ref{fig:generateLV}. 
The recovered pre-image returned by OASIS is visualized in Figure \ref{fig:resultLV}. 
For this problem, OASIS continues to maintain a high accuracy ($99.56\%$).

%and Table \ref{table:t4} summarizes the accuracy as well as sampling time for each point.  
 
%\begin{center}
%	\begin{table}	
%		\begin{tabular}{c c c c} 
%			\hline
%			  & \textbf{Accuracy} & \textbf{Training Time (s)} & \textbf{Prediction Time (s)} \\ 
%			Lotka-Volterra & 0.9976 & 307.08 & 17.05 \\
%			\hline
%		\end{tabular}
%		\caption{Result of Lotka-Volterra experiment}
%		\label{table:t4}
%	\end{table}
%\end{center}

%As a result, we gain an astonishing 0.9976 of accuracy and the outcome is almost impeccable compared to the ground truth, in which there are several flaws around the point $(0, 0)$. Considering the time, only 17 seconds is the duration for predicting more than 1000000 points and it takes more than 300 seconds for training our model.

\section{Discussions and conclusions}

We have proposed a novel active learning framework for the problem of set inversion and developed an efficient method to solve it. 
Our algorithm is implemented in a Python package OASIS.
The experimental results show that OASIS outperforms the current state-of-the-art program VSIVIA, especially in high-dimensional spaces. 
It is interesting to mention VSIVIA algorithm fails when the dimension is more than four while our algorithm is fairly robust to the increase of dimension.
The prediction time of OASIS is also noticeably faster than VSIVIA and has the potential to make significant improvement over VSIVIA in scenarios where prediction needs to be made as quickly as possible with limited computational power.
We provide an application of our method in studying the predator-prey problem in which we use the set inversion approach to recover set of the parameters of interest.

\section*{Acknowledgments}
BTN and DMN would like to thank The National Foundation for Science and Technology Development (NAFOSTED), University of Science, and Inspectorio Research Lab in Viet Nam for supporting two first authors throughout this paper. LSTH was supported by startup funds from Dalhousie University, the Canada Research Chairs program, and NSERC Discovery Grant.

\bibliographystyle{plain}
\bibliography{biblio}

\end{document}